\documentclass[runningheads]{llncs}

\usepackage[mobile]{eccv}

\usepackage{eccvabbrv}

\usepackage{graphicx}
\usepackage{booktabs}
\usepackage{multirow}
\usepackage[ruled]{algorithm2e} 
\usepackage{algorithmic}
\usepackage{float}
\usepackage[accsupp]{axessibility}  

\usepackage[pagebackref,breaklinks,colorlinks,citecolor=eccvblue]{hyperref}

\usepackage{orcidlink}

\begin{document}

\title{LightenDiffusion: Unsupervised Low-Light Image Enhancement with Latent-Retinex Diffusion Models} 

\titlerunning{LightenDiffusion}

\author{Hai Jiang\inst{1,5}\orcidlink{0000-0002-7087-6775} \and
Ao Luo\inst{2,5}\orcidlink{0000-0003-3494-8062} \and
Xiaohong Liu\inst{4}\orcidlink{0000-0001-6377-4730} \and \\
Songchen Han\inst{1}\orcidlink{0000-0001-8598-6558} \and
Shuaicheng Liu\inst{3,5,\dagger}\orcidlink{0000-0002-8815-5335}}

\authorrunning{Jiang et al.}

\institute{$^1$ Sichuan University, $^2$ Southwest Jiaotong University, \\ $^3$ University of Electronic Science and Technology of China, \\ $^4$ Shanghai Jiao Tong University, $^5$ Megvii Technology \\
\email{\{jianghai@stu.,hansongchen@\}scu.edu.cn, aoluo@swjtu.edu.cn, xiaohongliu@sjtu.edu.cn, liushuaicheng@uestc.edu.cn}\\
$^\dagger$ Corresponding Author}

\maketitle

\begin{abstract}
In this paper, we propose a diffusion-based unsupervised framework that incorporates physically explainable Retinex theory with diffusion models for low-light image enhancement, named LightenDiffusion. Specifically, we present a content-transfer decomposition network that performs Retinex decomposition within the latent space instead of image space as in previous approaches, enabling the encoded features of unpaired low-light and normal-light images to be decomposed into content-rich reflectance maps and content-free illumination maps. Subsequently, the reflectance map of the low-light image and the illumination map of the normal-light image are taken as input to the diffusion model for unsupervised restoration with the guidance of the low-light feature, where a self-constrained consistency loss is further proposed to eliminate the interference of normal-light content on the restored results to improve overall visual quality. Extensive experiments on publicly available real-world benchmarks show that the proposed LightenDiffusion outperforms state-of-the-art unsupervised competitors and is comparable to supervised methods while being more generalizable to various scenes. Our code is available at https://github.com/JianghaiSCU/LightenDiffusion.
\end{abstract}

\section{Introduction}\label{sec:introduction}
Images captured under weakly illuminated conditions suffer from various degradations such as poor visibility and noise, which leads to adverse impacts on the performance of downstream vision tasks~\cite{Classification, low_light_detection}. To transform low-light images into high-quality images, numerous works have been proposed in the past decades. Traditional methods~\cite{HE2, SRIE, HE1, NPE, LIME} mainly adopt hand-crafted priors, such as histogram equalization (HE)~\cite{HE} and Retinex theory~\cite{Retinex}, to improve contrast and restore details. However, it is difficult to adopt a suitable prior for various illumination conditions since low-light image enhancement (LLIE) is an ill-posed problem, thus limiting the practical application of these methods.

\begin{figure}[!t]
    \centering
    \includegraphics[width=0.68\linewidth]{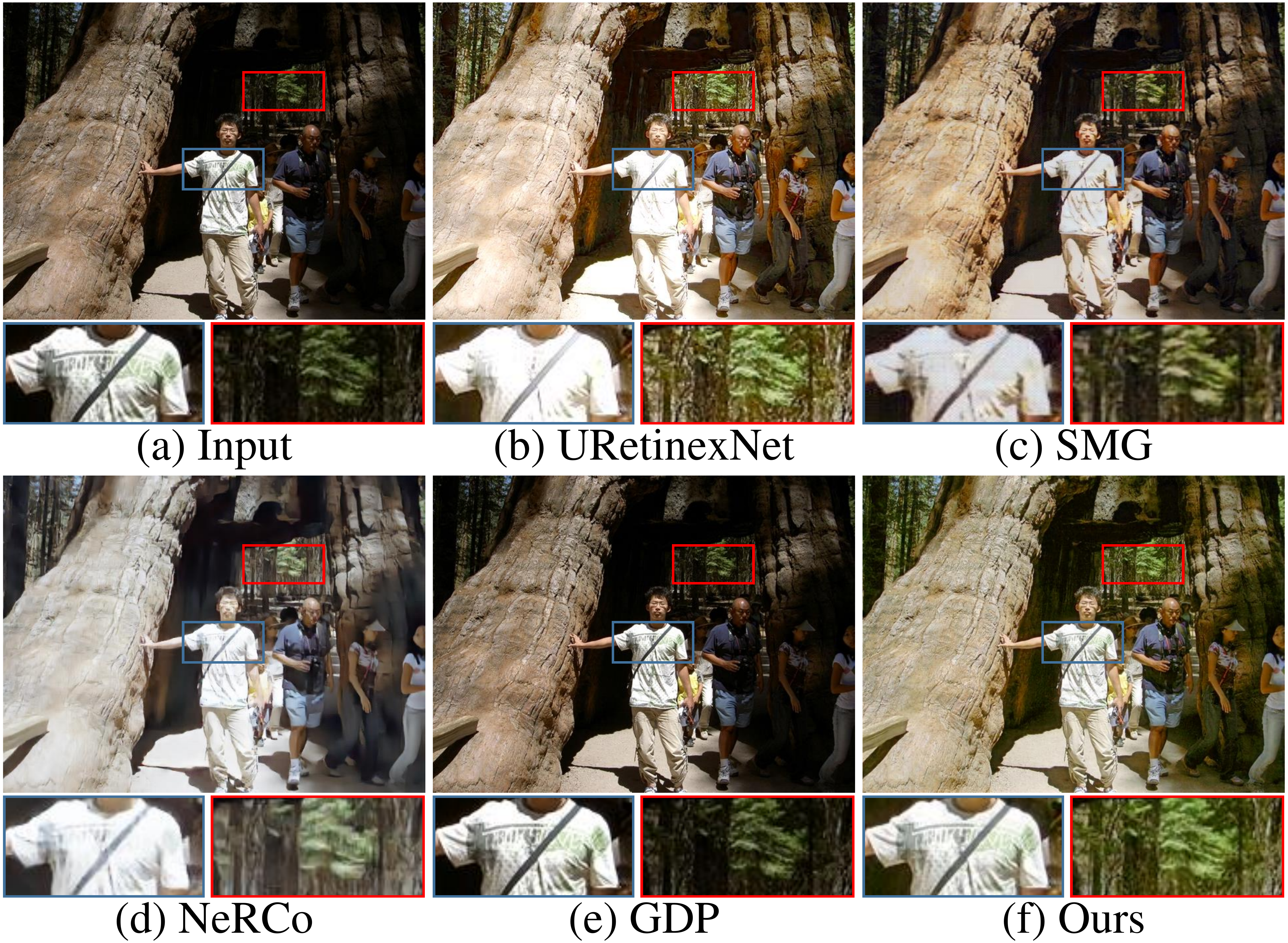}
    \caption{Visual comparisons of our method with recent state-of-the-art supervised and unsupervised LLIE methods UReinexNet~\cite{Uretinex-net}, SMG~\cite{SMG}, NeRCo~\cite{NeRCo}, and GDP~\cite{GDP}. Previous methods appear incorrect exposure, color distortion, blurred details, or noise amplification to degrade visual quality, while our method properly improves global and local contrast, presents a vivid color, and avoids introducing artifacts.}
    \label{fig: teaser}
\end{figure}

These issues have been partially resolved with the development of deep learning, where learning-based methods~\cite{LLNet, RetinexNet, Zero-DCE, RUAS, SNRNet, Bread, SMG, Semantic_Aware, PairLIE, GCCIM, Retinexformer} can directly learn the mapping from low-light images to normal-light images through powerful network architectures and sophisticated learning strategies, which present more robustness than traditional methods. While learning-based methods achieve remarkable progress in LLIE, they often suffer from the overfitting problem and struggle with poor generalization ability, resulting in outcomes with unsatisfactory visual fidelity. As shown in Fig.~\ref{fig: teaser}(b)-(d), previous state-of-the-art supervised methods URetinexNet~\cite{Uretinex-net} and SMG~\cite{SMG}, as well as unsupervised method NeRCo~\cite{NeRCo} present incorrect overexposure, color distortion, blurred details or noise amplification in the highlighted regions.

Recently, generative model-based methods~\cite{EnlightenGAN, low_light_GAN, LUD_VAE} have emerged for LLIE as promising approaches to obtain better perceptual quality, in which diffusion models~\cite{ddpm, ddim} have gained attention for their impressive generative ability and being free from instability and mode-collapse problems present in previous generative models such as generative adversarial networks (GANs) and variational autoencoders (VAEs). Most diffusion-based methods~\cite{SR3, palette, weatherdiff, PyDiff, DiffLL, Diff-Retinex, GSAD} utilize large-scale paired data with conditional mechanism~\cite{conditional_ddpm} for supervised learning, which enable favorable contrast enhancement and details reconstruction, while it is challenging to collect paired distorted/sharp images in the real world. To leverage the label-free characteristic of unsupervised learning to improve the generalization of diffusion models, some methods~\cite{DDNM, ddrm, Repaint, DiffPIR, GDP} employ zero-shot solutions that utilize well-established priors from pre-trained diffusion models for restoration without training from scratch. However, these methods are limited by the known degradation modes and thus tend to perform poorly in real-world scenes where distortions are diverse and unknown. As shown in Fig.~\ref{fig: teaser}(e), the zero-shot-based method GDP~\cite{GDP} produces an under-enhancement result.

To this end, we propose a diffusion-based learnable unsupervised framework, dubbed LightenDiffusion, which incorporates physically interpretable Retinex theory with diffusion models to learn degradation modes of various scenes. It accomplishes this by training on extensive unpaired real-world data, ultimately achieving visually favorable LLIE. Specifically, we first convert the unpaired low-light and normal-light images into latent space, where the encoded features are decomposed into content-rich reflectance maps that contain abundant content-related details and content-free illumination maps that only represent the lighting conditions through the proposed content-transfer decomposition network. Subsequently, the reflectance map of the low-light feature and the illumination map of the normal-light feature serve as input to the diffusion model for restoration with the guidance of the low-light feature. Moreover, the distribution learned by the diffusion model may be disrupted once the estimated normal-light illumination map still preserves certain content information, leading to the restored result being interfered by the normal-light image content. Therefore, we propose a self-constrained consistency loss to promote the diffusion model to reconstruct images with the same intrinsic content information as input low-light images. As shown in Fig.~\ref{fig: teaser}(f), our method properly improves global and local contrast, prevents overcorrection on the well-exposed region, and avoids artifacts or noise amplification. Extensive experiments show that our method outperforms existing state-of-the-art competitors quantitatively and visually. The application for low-light face detection also reveals the potential practical values of our method. 

Our contributions can be summarized as follows:
\begin{itemize}
    \item We propose a diffusion-based framework, termed LightenDiffusion, that leverages the advantages of Retinex theory and the generative ability of diffusion models for unsupervised low-light image enhancement, with a self-constrained consistency loss further proposed to improve visual quality.
    \item We propose a content-transfer decomposition network that performs decomposition in the latent space, aiming to obtain content-rich reflectance maps and content-free illumination maps to promote unsupervised restoration.
    \item Extensive experiments demonstrate that our method outperforms existing state-of-the-art unsupervised competitors while being comparable and having better generalization abilities than supervised methods.
\end{itemize} 

\section{Related Work}\label{sec:related_work}

\subsection{Low-Light Image Enhancement}\label{subsec:low_light_image_enhancement}
Numerous works have been proposed to transform poorly illuminated images into visually pleasant normal-light images. Traditional methods depend on hand-crafted optimization rules such as Histogram Equalization (HE)~\cite{HE} and Retinex theory~\cite{Retinex}. HE-based methods~\cite{HE1, HE2} aim to change the histogram distribution of the image to improve the contrast. Retinex-based methods~\cite{SRIE, LIME} first decompose an image into a reflectance map and an illumination map, with the visual quality being improved by changing the dynamic range of the illumination map.

Recently, learning-based methods have achieved remarkable results in the LLIE task and show more robustness than traditional methods, which can be mainly categorized as supervised, semi-supervised, and unsupervised. The former~\cite{LLNet, MIRNet, SNRNet, Bread, SMG, Semantic_Aware, GCCIM, RFLLIE} leverage powerful network architectures to learn mappings from low-light images to normal-light ones in an end-to-end manner. Some approaches~\cite{RetinexNet, KinD++, Advanced_RetinexNet, Uretinex-net, Retinexformer} combine Retinex theory with deep networks to establish learnable decomposition and adjustment frameworks. However, supervised methods rely on large-scale paired datasets for training and thus suffer from limited generalization ability. To address these issues, unsupervised methods~\cite{Zero-DCE, EnlightenGAN, RUAS, SCI, PairLIE, NeRCo} utilize their characteristics of not requiring paired data to solve the LLIE by employing adversarial learning, curve estimation, or neural architecture search with better generalization in real-world scenes. Semi-supervised methods~\cite{DRBN, semi_cvpr23} combine the advantages of supervised and unsupervised learning to achieve stable training while maintaining better generalization capability.

\subsection{Diffusion-based Image Restoration}\label{subsec:diffusion_based_image_restoration}
With the development of diffusion models (DMs) in low-level vision~\cite{ddpm, ddim, latent_diffusion, DMHomo, RecDiffusion, FlowDiffuser, HandBooster, rolling_shutter}, many works have been conducted to explore their performance in image restoration tasks, such as super-resolution~\cite{SR3, IDM}, inpainting~\cite{palette, Smartbrush}, weather removal~\cite{weatherdiff, IR_SDE, Refusion}, and low-light image enhancement~\cite{PyDiff, Diff-Retinex, Exposurediffusion, DiffLL, GSAD, Retidiff}. Most methods utilize the conditional mechanism~\cite{conditional_ddpm} to train diffusion models from scratch with paired data, where degraded images serve as guidance in the diffusion processes. In contrast, some methods~\cite{ddrm, Repaint, GDP, DDNM, DiffPIR} employ zero-shot strategies using pre-trained diffusion models to restore degraded images without reference images directly. They leverage the priors from the pre-trained models for restoration, rather than deriving the capability from the training datasets. Although zero-shot approaches provide an attractive alternative, their performance is hampered by the pre-trained models, leading to the restored results with unsatisfactory visual quality. In this paper, we propose to incorporate the physically explainable Retinex theory with diffusion models to achieve visually satisfactory LLIE in an unsupervised manner.
\begin{figure}[!t]
    \centering
    \includegraphics[width=\linewidth]{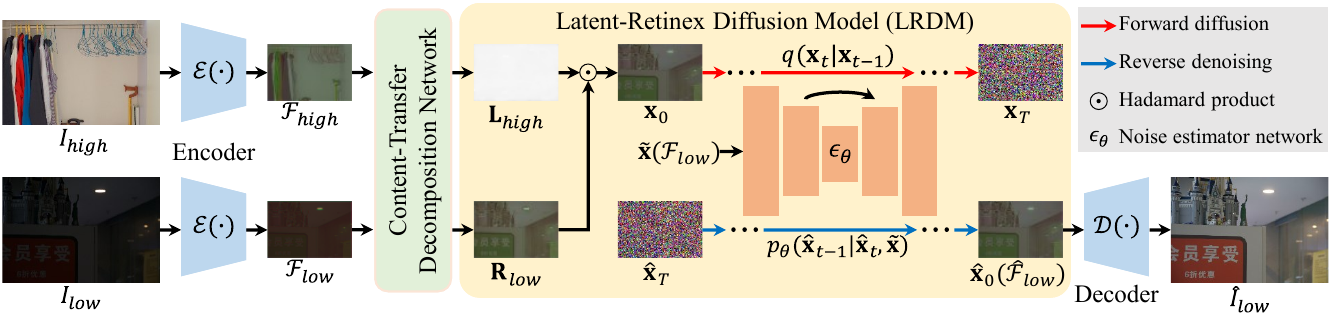}
    \caption{The overall pipeline of our proposed framework. We first employ an encoder $\mathcal{E}(\cdot)$ to convert the unpaired low-light image $I_{low}$ and normal-light image $I_{high}$ into latent space denoted as $\mathcal{F}_{low}$ and $\mathcal{F}_{high}$. The encoded features are sent to the proposed content-transfer decomposition network (CTDN) to generate content-rich reflectance maps denoted as $\mathbf{R}_{low}$ and $\mathbf{R}_{high}$ and content-free illumination maps as $\mathbf{L}_{low}$ and $\mathbf{L}_{high}$. Then, the reflectance map of the low-light image $\mathbf{R}_{low}$ and the illumination of the normal-light image $\mathbf{L}_{high}$ are taken as the input of the diffusion model to perform the forward diffusion process. Finally, we perform the reverse denoising process to gradually transform the randomly sampled Gaussian noise $\hat{\mathbf{x}}_{T}$ into the restored feature $\mathcal{\hat{F}}_{low}$ with the guidance of the low-light feature $\mathcal{F}_{low}$ denoted as $\tilde{\mathbf{x}}$, and subsequently send it to a decoder $\mathcal{D}(\cdot)$ to produce the final result $\hat{I}_{low}$.}
    \label{fig: Pipeline}
\end{figure}

\section{Methodology}\label{sec:method}
\subsection{Overview}\label{subsec:overview}
The overall pipeline of our proposed framework is illustrated in Fig.~\ref{fig: Pipeline}. Given an unpaired low-light image $I_{low} \in \mathbb{R}^{H\times W \times 3}$ and normal-light image $I_{high} \in \mathbb{R}^{H\times W \times 3}$, we first employ an encoder $\mathcal{E}(\cdot)$, which consists of $k$ cascaded residual blocks where each block downsamples the input by a scale of 2 using a max-pooling layer, to transform the input images into latent space denoted as $\mathcal{F}_{low} \in \mathbb{R}^{\frac{H}{2^{k}} \times \frac{W}{2^{k}} \times C}$ and $\mathcal{F}_{high} \in \mathbb{R}^{\frac{H}{2^{k}} \times \frac{W}{2^{k}} \times C}$. Then, we design a content-transfer decomposition network (CTDN) to decompose the features into content-rich reflectance maps $\mathbf{R}_{low}$ and $\mathbf{R}_{high}$ and content-free illumination maps $\mathbf{L}_{low}$ and $\mathbf{L}_{high}$. Subsequently, the $\mathbf{R}_{low}$ and the $\mathbf{L}_{high}$ serve as input for the diffusion model with the guidance of the low-light feature to generate the restored feature $\hat{\mathcal{F}}_{low}$. Finally, the restored feature will be sent to a decoder $\mathcal{D}(\cdot)$ for reconstruction to produce the final restored image $\hat{I}_{low}$.

\subsection{Content-Transfer Decomposition Network}\label{subsec:CTDN}
\begin{figure}[!t]
    \centering
    \includegraphics[width=0.8\linewidth]{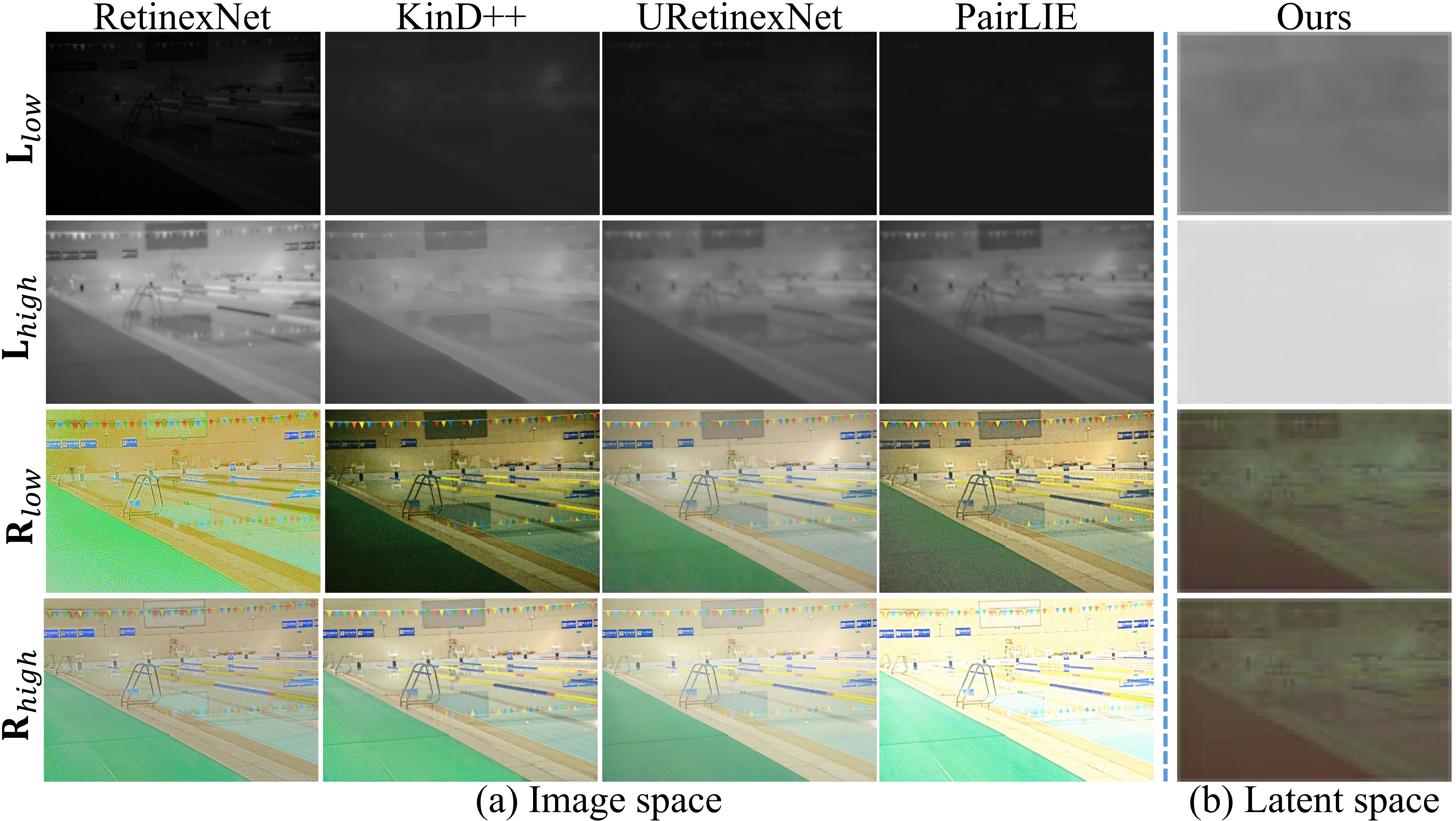}
    \caption{Illustration of the decomposition results obtained by different methods. (a) shows the results of previous methods, i.e., RetinexNet~\cite{RetinexNet}, KinD++~\cite{KinD++}, URetinexNet~\cite{Uretinex-net}, and PairLIE~\cite{PairLIE}, that perform decomposition in image space. (b) presents the results of our CTDN that performs decomposition in latent space. Our method can generate content-rich reflectance maps and content-free illumination maps.}
    \label{fig: Retinex_comapre}
\end{figure}

The Retinex theory~\cite{Retinex} assumes that an image $I$ can be decomposed into a reflectance map $\mathbf{R}$ and an illumination map $\mathbf{L}$ as:
\begin{equation}\label{eq:1}
    I = \mathbf{R} \odot \mathbf{L}
\end{equation}
where $\odot$ denotes Hadamard product operation. $\mathbf{R}$ represents the inherent content information that should be consistent under diverse illumination conditions, while $\mathbf{L}$ indicates the contrast and brightness information that should be local smoothness. However, existing methods~\cite{RetinexNet, KinD++, Uretinex-net, Retinexformer, Diff-Retinex, PairLIE} typically perform decomposition in the image space to obtain the above components, which results in the content information not being fully decomposed into the reflectance map and partially retained in the illumination map, as shown in Fig.~\ref{fig: Retinex_comapre}(a). 

\begin{figure}[!t]
    \centering
    \includegraphics[width=0.8\linewidth]{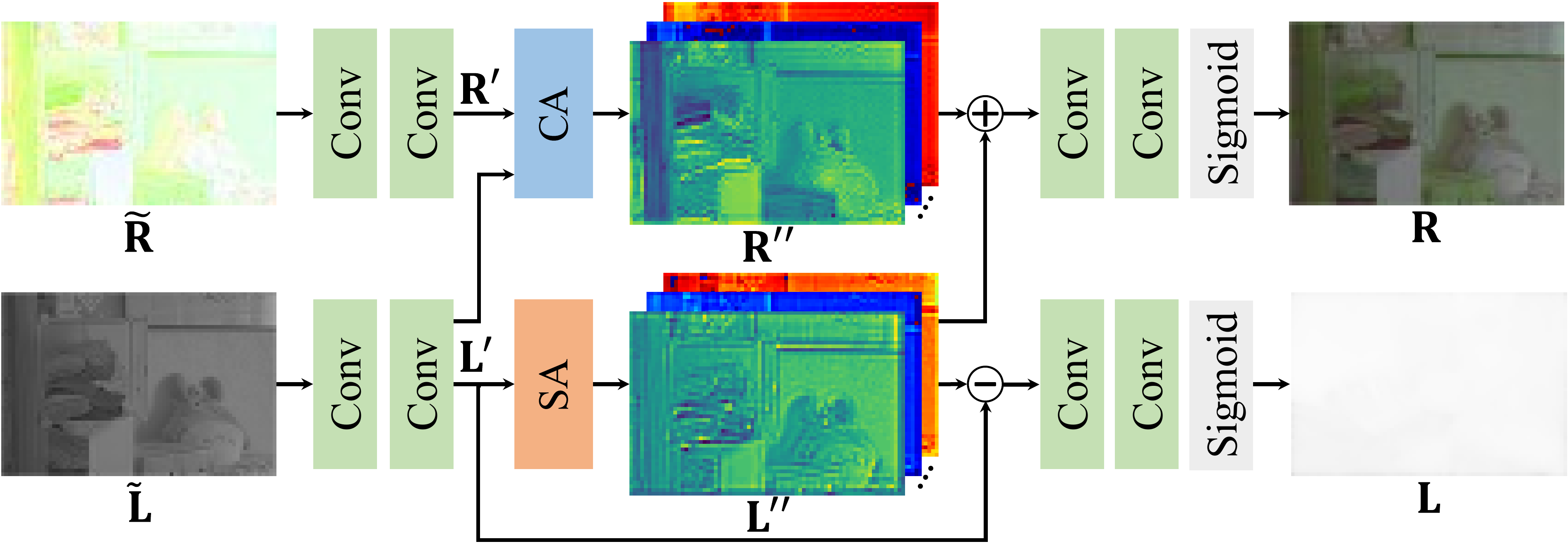}
    \caption{The detailed architecture of our proposed CTDN.}
    \label{fig: CTDN}
\end{figure}

To alleviate this issue, we introduce a content-transfer decomposition network (CTDN) that performs decomposition within the latent space. By encoding the content information in this latent space, the CTDN facilitates the generation of reflectance maps containing abundant content-related details and illumination maps that remain unaffected by content-related influences. As shown in Fig.~\ref{fig: CTDN}, we first estimate the initial reflectance and illumination maps following~\cite{LIME} as:
\begin{equation}\label{eq:2}
    \tilde{\mathbf{L}}(x) = \max_{c \in [0, C)}\mathcal{F}^{c}(x), \tilde{\mathbf{R}}(x) = \mathcal{F}(x) / (\tilde{\mathbf{L}}(x)+\tau),
\end{equation}
for each pixel $x$, where $\tau$ is a small constant to avoid zero denominator. The estimated maps are refined through two branches, in which we first employ several convolutional blocks to obtain the embedded features as $\mathbf{L}^{'} = \operatorname{Convs}(\tilde{\mathbf{L}}), \mathbf{R}^{'} = \operatorname{Convs}(\tilde{\mathbf{R}})$. Subsequently, we utilize a cross-attention (CA)~\cite{cross_attention} module to leverage the illumination map to reinforce the content information in the reflectance map as $\mathbf{R}^{''} = \operatorname{CA}(\mathbf{R}^{'}, \mathbf{L}^{'})$. Moreover, a self-attention module (SA)~\cite{self_attention} is adopted to further extract content information in the illumination map, denoted as $\mathbf{L}^{''} = \operatorname{SA}(\mathbf{L}^{'})$, and complement it to the reflectance map. The final output reflectance map $\mathbf{R}$ and illumination map $\mathbf{L}$ can be expressed as $\mathbf{R} = \operatorname{Convs}(\mathbf{R}^{''} + \mathbf{L}^{''})$ and $\mathbf{L} = \operatorname{Convs}(\mathbf{L}^{'} - \mathbf{L}^{''})$. As shown in Fig.~\ref{fig: Retinex_comapre}(b), our CTDN can generate content-rich reflectance maps that fully represent the intrinsic information of the image, and content-free illumination maps that only reveal the lighting conditions.

\subsection{Latent-Retinex Diffusion Models}\label{subsec:LRDM}
One straightforward way to obtain the enhanced feature in the ideal case is to multiply the reflectance map of the low-light feature with the illumination map of the normal-light image as $\hat{\mathcal{F}}_{low} = \mathbf{R}_{low} \odot \mathbf{L}_{high}$. However, there are two challenges with the above approach: 1) Retinex decomposition inevitably encounters information loss; 2) the restored image would present artifacts once the illumination map of the reference normal-light image still contains stubborn content information. Although our CTDN is generally effective in most scenes, there may be challenging cases where the accuracy of the estimated illumination map is compromised. To address these problems, we propose a Latent-Retinex diffusion model (LRDM) that leverages the generative ability of diffusion models to compensate for content loss and eliminate potential unexpected artifacts. Our approach follows standard diffusion models~\cite{ddpm, conditional_ddpm, ddim} that perform forward diffusion and reverse denoising processes to generate restored results. 

\textbf{Forward Diffusion.} Given the decomposition components of unpaired images, we take the reflectance map of the low-light image $\mathbf{R}_{low}$ and the illumination map of the normal-light image $\mathbf{L}_{high}$ as input, denoted as $\mathbf{x}_{0} = \mathbf{R}_{low} \odot \mathbf{L}_{high}$, to perform the forward diffusion process, and uses a pre-defined variance schedule $\{\beta_{1}, \beta_{2}, \dots, \beta_{T}\}$ to progressively transform $\mathbf{x}_{0}$ into Gaussian noise $\mathbf{x}_{T} \sim \mathcal{N}(\mathbf{0},\mathbf{I})$ through $T$ steps, which can be formulated as:
\begin{equation}\label{eq:3}
    q(\mathbf{x}_t \mid \mathbf{x}_{t-1})=\mathcal{N}(\mathbf{x}_t ; \sqrt{1-\beta_t} \mathbf{x}_{t-1}, \beta_t \mathbf{I}),
\end{equation}
where $\mathbf{x}_t$ indicates the noisy data at time-step $t \in [0, T]$. By utilizing parameter renormalization, we can merge and refine multiple Gaussian distributions to obtain the $\mathbf{x}_t$ directly from the input $\mathbf{x}_0$ and simplify Eq.(\ref{eq:4}) into a closed form as $\mathbf{x}_t=\sqrt{\bar{\alpha}_t} \mathbf{x}_0+\sqrt{1-\bar{\alpha}_t} \boldsymbol{\epsilon}_t$, where $\alpha_t = 1 - \beta_t$, $\bar{\alpha}_t=\prod_{i=0}^t \alpha_i$, and $\boldsymbol{\epsilon}_t \sim \mathcal{N}(\mathbf{0},\mathbf{I})$.

\textbf{Reverse Denoising.} By utilizing the editing and data synthesis capabilities offered by conditional diffusion models~\cite{conditional_ddpm}, we aim to gradually denoise a randomly sampled Gaussian noise $\hat{\mathbf{x}}_{T} \sim \mathcal{N}(\mathbf{0},\mathbf{I})$ into a sharp result $\hat{\mathbf{x}}_{0}$ with the guidance of the encoded feature $\mathcal{F}_{low}=\mathcal{E}(I_{low})$ of the low-light image denoted as $\tilde{\mathbf{x}}$, which facilitates resulting in high fidelity of restored results to the distribution conditioned on $\tilde{\mathbf{x}}$. The reverse denoising process can be formulated as:
\begin{equation}\label{eq:4}
    p_\theta(\mathbf{\hat{x}}_{t-1} \mid \mathbf{\hat{x}}_t, \tilde{\mathbf{x}})=\mathcal{N}(\mathbf{\hat{x}}_{t-1};\boldsymbol{\mu}_\theta(\mathbf{\hat{x}}_t, \tilde{\mathbf{x}}, t), \sigma_t^2 \mathbf{I}),
\end{equation}
where $\sigma_t^2=\frac{1-\bar{\alpha}_{t-1}}{1-\bar{\alpha}_t} \beta_t$ is the variance and $\boldsymbol{\mu}_\theta(\mathbf{\hat{x}}_t, \tilde{\mathbf{x}}, t)=\frac{1}{\sqrt{\alpha_t}}(\mathbf{\hat{x}}_t-\frac{\beta_t}{\sqrt{1-\bar{\alpha}_t}} \boldsymbol{\epsilon}_\theta(\mathbf{\hat{x}}_t, \tilde{\mathbf{x}}, t))$ is the mean value. 
\begin{algorithm}[!t]
    \caption{LRDM training}
    \label{algo:1} 
    \SetKwData{Left}{left}\SetKwData{This}{this}\SetKwData{Up}{up} \SetKwFunction{Union}{Union}\SetKwFunction{FindCompress}{FindCompress} \SetKwInOut{Input}{input}\SetKwInOut{Output}{output}
    
    \Input{The decomposition results $\mathbf{R}_{low}$ and $\mathbf{L}_{high}$, low-light feature $\mathcal{F}_{low}$, time step $T$, and sampling step $S$.} 

    $\mathbf{x}_{0} = \mathbf{R}_{low} \odot \mathbf{L}_{high}$, $\tilde{\mathbf{x}} = \mathcal{F}_{low}$
    
    \While{Not converged}{
    
    $\epsilon_t \sim \mathcal{N}(\mathbf{0},\mathbf{I})$, $t \sim \operatorname{Uniform}\{1,\cdots,T\}$
    
    Perform gradient descent steps on
    $\nabla_\theta\|\boldsymbol{\epsilon}_t-\boldsymbol{\epsilon}_\theta(\sqrt{\bar{\alpha}_t} \mathbf{x}_0+\sqrt{1-\bar{\alpha}_t} \boldsymbol{\epsilon}_t, \tilde{\mathbf{x}}, t)\|^2$
    \par
    $\mathbf{\hat{x}}_T \sim \mathcal{N}(\mathbf{0},\mathbf{I})$
    
    \For{$i = S:1$}{ 
    $t = (i-1) \cdot T/S + 1$
    
    $t_{\operatorname{next}} = (i-2) \cdot T/S + 1$ if $i > 1$, else $0$

    $\mathbf{\hat{x}}_{t} \leftarrow \sqrt{\bar{\alpha}_{t_{\operatorname{next}}}}(\frac{\mathbf{\hat{x}}_t-\sqrt{1-\bar{\alpha}_t} \cdot \epsilon_\theta(\mathbf{\hat{x}}_t, \tilde{\mathbf{x}}, t)}{\sqrt{\bar{\alpha}_t}}) +\sqrt{1-\bar{\alpha}_{t_{\operatorname{next}}}} \cdot \epsilon_\theta(\mathbf{\hat{x}}_t, \tilde{\mathbf{x}}, t)$
    }
    
    Perform gradient descent steps on
    $\nabla_\theta\|\mathbf{R}_{low} \odot \mathbf{L}_{low}^{\gamma} - \mathbf{\hat{x}}_0\|^2$}
     
    \Output{$\theta$}
\end{algorithm}

In the training phase, the objective of the diffusion model is to optimize the parameters $\theta$ of the network $\boldsymbol{\epsilon}_\theta$ to promote the estimated noise vector $\boldsymbol{\epsilon}_\theta(\mathbf{x}_t, \tilde{\mathbf{x}}, t)$ close to Gaussian noise like~\cite{ddpm}, which is formulated as:
\begin{equation}\label{eq:5}
    \mathcal{L}_{diff} = \|\boldsymbol{\epsilon}_t-\boldsymbol{\epsilon}_\theta(\mathbf{x}_t, \tilde{\mathbf{x}}, t)\|_{2}.
\end{equation}
During inference, we obtain the restored feature $\mathcal{\hat{F}}_{low}$ from the distribution learned by the diffusion model through reverse denoising process with implicit sampling strategy~\cite{ddim}, and subsequently send it to the decoder to produce the final result $\hat{I}_{low}$. However, as mentioned above, the input $\mathbf{x}_{0}$ would present artifacts once the estimated illumination map still contains content information, which may affect the learned distribution and result in the $\mathcal{\hat{F}}_{low}$ being disrupted.

Therefore, we propose a self-constrained consistency loss $\mathcal{L}_{scc}$ to enable the restored feature to share the same intrinsic information as the input low-light image. Specifically, we first perform the reverse denoising process in the training phase following~\cite{PyDiff, DiffLL, GSAD} to generate the restored feature and construct a pseudo label $\mathcal{\tilde{F}}_{low}$ from decomposition results of the low-light image as a reference based on traditional Gamma correction approaches as $\mathcal{\tilde{F}}_{low} = \mathbf{R}_{low} \odot \mathbf{L}_{low}^{\gamma}$, where $\gamma$ is the illumination correction factor. Thus, the $\mathcal{L}_{scc}$ aims to constrain the feature similarity to prompt the diffusion model to reconstruct $\hat{I}_{low}$ as:
\begin{equation}\label{eq:6}
    \mathcal{L}_{scc} = \|\mathcal{\tilde{F}}_{low} - \mathcal{\hat{F}}_{low}\|_{1}.
\end{equation}
Overall, the training strategy of our LRDM is summarized in Alg.~\ref{algo:1} and the objective function used for optimization is rewritten as $\mathcal{L} = \mathcal{L}_{diff} + \lambda_1 \mathcal{L}_{scc}$.

\subsection{Network Training}\label{subsec:network_training}
Our approach adopts a two-stage strategy for network training. In the first stage, we follow~\cite{PairLIE} that utilizes paired low-quality images, denoted as $I_{low}^{1}$ and $I_{low}^{2}$, from the SICE dataset~\cite{SICE} to optimize the encoder $\mathcal{E}(\cdot)$, CTDN, and decoder $\mathcal{D}(\cdot)$, while freezing the parameters of the diffusion model. The encoder and decoder are optimized with the content loss $\mathcal{L}_{con}$ as:
\begin{equation}\label{eq:7}
    \mathcal{L}_{con} = \sum_{i=1}^2\|I_{low}^{i} - \mathcal{D}(\mathcal{E}(I_{low}^{i}))\|_{2}.
\end{equation}

The CTDN is optimized with the decomposition loss $\mathcal{L}_{dec}$ as~\cite{RetinexNet, KinD++, Diff-Retinex} that consist of the reconstruction loss $\mathcal{L}_{rec}$, the reflectance consistency loss $\mathcal{L}_{ref}$, and the illumination smoothing loss $\mathcal{L}_{ill}$. The $\mathcal{L}_{rec}$ aims to guarantee the decomposed components can reconstruct the input features, which is expressed as:
\begin{equation}\label{eq:8}
    \mathcal{L}_{rec} = \sum_{i=1}^2\sum_{j=1}^2\|\mathcal{F}_{low}^{j} - \mathbf{R}_{low}^{i} \odot \mathbf{L}_{low}^{j} \|_{1}.
\end{equation}
The $\mathcal{L}_{ref}$ aims to enforce the network to produce invariant reflectance maps and the $\mathcal{L}_{ill}$ is adopted to guarantee the illumination map to be local smoothness, which can be expressed respectively as:
\begin{equation}\label{eq:9}
    \mathcal{L}_{ref} = \|\mathbf{R}_{low}^{1} - \mathbf{R}_{low}^{2}\|_{1}, \mathcal{L}_{ill} = \sum_{i=1}^2\|\nabla \mathbf{L}_{low}^{i} \cdot \exp(-\lambda_g \nabla \mathbf{R}_{low}^{i})\|_{2},
\end{equation}
where $\nabla$ denotes the horizontal and vertical gradients, and $\lambda_g$ is the coefficient to balance the perceived strength of the structure. The overall decomposition loss used to optimize the CTDN is formulated as $\mathcal{L}_{dec} = \mathcal{L}_{rec} + \lambda_{2}\mathcal{L}_{ref} + \lambda_{3}\mathcal{L}_{ill}$.

In the second stage, we collect $\sim$180k unpaired low/normal-light image pairs to optimize the diffusion model while freezing the parameters of other modules.

\section{Experiments}\label{sec:experiment}
\subsection{Experimental Settings}\label{subsec:experimental_settings}
\textbf{Implementation Details.} We implement the proposed method with PyTorch on four NVIDIA RTX 2080Ti GPUs, where the batch size and patch size are set to 12 and 512 $\times$512. The networks can be converged after training in two stages with $1 \times 10^5$ and $4 \times 10^5$ iterations, respectively. We employ the Adam optimizer~\cite{Adam} for optimization with the initial learning rate set to $1 \times 10^{-4}$ in the first stage and decays by a factor of 0.8 while reinitializing it to a fixed value of $2 \times 10^{-5}$ in the second stage. The feature downsampling scale $k$ and the illumination correction factor $\gamma$ are set to 3 and 0.2, respectively. The hyper-parameters $\lambda_{1}$, $\lambda_{2}$, $\lambda_{3}$, and $\lambda_{g}$ are empirically set to 0.01, 0.1, 0.01, and 10, respectively. For our LRDM, the U-Net~\cite{Unet} architecture is adopted as the noise estimator network with the time step $T$ and sampling step $S$ set to 1000 and 20 for the forward diffusion and reverse denoising processes, respectively.
\begin{table}[!t]
   \centering
    \caption{Quantitative comparisons on the paired LOL~\cite{RetinexNet} and LSRW~\cite{R2RNet} datasets, and unpaired DICM~\cite{DICM}, NPE~\cite{NPE}, and VV~\cite{VV} datasets. The best results are highlighted in \textbf{bold}. `T', `SL', `SSL', and `UL' indicate that the methods belong to traditional, supervised, semi-supervised, and unsupervised methods, respectively.}
    \resizebox{\linewidth}{!}{
    \begin{tabular}{c|l|ccc|ccc|cc|cc|cc}
    \toprule
    \multirow{2}[4]{*}{Type} & \multirow{2}[4]{*}{Method} & \multicolumn{3}{c|}{LOL~\cite{RetinexNet}} & \multicolumn{3}{c|}{LSRW~\cite{R2RNet}} & \multicolumn{2}{c|}{DICM~\cite{DICM}} & \multicolumn{2}{c|}{NPE~\cite{NPE}} & \multicolumn{2}{c}{VV~\cite{VV}} \\
    \cmidrule{3-14} & & PSNR $\uparrow$ & SSIM $\uparrow$ & LPIPS $\downarrow$ & PSNR $\uparrow$ & SSIM $\uparrow$ & LPIPS $\downarrow$ & NIQE $\downarrow$ & PI $\downarrow$ & NIQE $\downarrow$ & PI $\downarrow$ & NIQE$\downarrow$ & PI $\downarrow$ \\
    \midrule
    \multirow{4}[2]{*}{T} 
    & LIME~\cite{LIME} & 17.546 & 0.531 & 0.290 & 17.342 & 0.520 & 0.416 & 4.476 & 4.216 & 4.170 & 3.789 & 3.713 & 3.335 \\
    & SDDLLE~\cite{SDDLLE} & 13.342 & 0.634 & 0.261 & 14.708 & 0.486 & 0.382 & 4.581 & 3.828 & 4.179 & 3.315 & 4.274 & 3.382 \\
    & CDEF~\cite{CDEF}  & 16.335 & 0.585 & 0.351 & 16.758 & 0.465 & 0.314 & 4.142 & 4.242 & 3.862 & 2.910 & 5.051 & 3.272 \\
    & BrainRetinex~\cite{BrainRetinex} & 11.063 & 0.475 & 0.327 & 12.506 & 0.390 & 0.374 & 4.350 & 3.555 & 3.707 & 3.044 & 4.031 & 3.114 \\
    \midrule
    \multirow{7}[2]{*}{SL} 
    & RetinexNet~\cite{RetinexNet} & 16.774 & 0.462 & 0.390 & 15.609 & 0.414 & 0.393 & 4.487 & 3.242 & 4.732 & 3.219 & 5.881 & 3.727 \\
    & KinD++~\cite{KinD++} & 17.752 & 0.758 & 0.198 & 16.085 & 0.394 & 0.366 & 4.027 & 3.399 & 4.005 & 3.144 & 3.586 & 2.773 \\
    & LCDPNet~\cite{LCDPNet} & 14.506 & 0.575 & 0.312 & 15.689 & 0.474 & 0.344 & 4.110 & 3.250 & 4.106 & 3.127 & 5.039 & 3.347 \\
    & URetinexNet~\cite{Uretinex-net} & 19.842 & 0.824 & 0.128 & 18.271 & 0.518 & 0.295 & 4.774 & 3.565 & 4.028 & 3.153 & 3.851 & 2.891 \\
    & SMG~\cite{SMG} & \textbf{23.814} & 0.809 & 0.144 & 17.579 & 0.538 & 0.456 & 6.224 & 4.228 & 5.300 & 3.627 & 5.752 & 3.757 \\
    & PyDiff~\cite{PyDiff} &  23.275 & \textbf{0.859} & \textbf{0.108} & 17.264 & 0.510 & 0.335 & 4.499 & 3.792 & 4.082 & 3.268 & 4.360 & 3.678 \\
    & GSAD~\cite{GSAD} &  22.021 & 0.848 & 0.137 & 17.414 & 0.507 & \textbf{0.294} & 4.496 & 3.593 & 4.489 & 3.361 & 5.252 & 3.657 \\
    \midrule
    \multirow{2}[2]{*}{SSL}    
    & DRBN~\cite{DRBN} & 16.677 & 0.730 & 0.252 & 16.734 & 0.507 & 0.376 & 4.369 & 3.800 & 3.921 & 3.267 & 3.671 & 3.117 \\
    & BL~\cite{BL} & 10.305 & 0.401 & 0.382 & 12.444 & 0.333 & 0.384 & 5.046 & 4.055 & 4.885 & 3.870 & 5.740 & 4.030 \\
    \midrule
    \multirow{8}[2]{*}{UL} 
    & Zero-DCE~\cite{Zero-DCE} & 14.861 & 0.562 & 0.330 & 15.867 & 0.443 & 0.315 & 3.951 & 3.149 & 3.826 & 2.918 & 5.080 & 3.307 \\
    & EnlightenGAN~\cite{EnlightenGAN} & 17.606 & 0.653 & 0.319 & 17.106 & 0.463 & 0.322 & 3.832 & 3.256 & 3.775 & 2.953 & 3.689 & 2.749 \\
    & RUAS~\cite{RUAS} & 16.405 & 0.503 & 0.257 & 14.271 & 0.461 & 0.455 & 7.306 & 5.700 & 7.198 & 5.651 & 4.987 & 4.329 \\
    & SCI~\cite{SCI} & 14.784 & 0.525 & 0.333 & 15.242 & 0.419 & 0.321 & 4.519 & 3.700 & 4.124 & 3.534 & 5.312 & 3.648 \\
    & GDP~\cite{GDP} & 15.896 & 0.542 & 0.337 & 12.887 & 0.362 & 0.386 & 4.358 & 3.552 & 4.032 & 3.097 & 4.683 & 3.431 \\
    & PairLIE~\cite{PairLIE} & 19.514 & 0.731 & 0.254 & 17.602 & 0.501 & 0.323 & 4.282 & 3.469 & 4.661 & 3.543 & 3.373 & 2.734 \\
    & NeRCo~\cite{NeRCo} & 19.738 & 0.740 & 0.239 & 17.844 & 0.535 & 0.371 & 4.107 & 3.345 & 3.902 & 3.037 & 3.765 & 3.094 \\
    & Ours & 20.453 & 0.803 & 0.192 & \textbf{18.555} & \textbf{0.539} & 0.311 & \textbf{3.724} & \textbf{3.144} & \textbf{3.618} & \textbf{2.879} & \textbf{2.941} & \textbf{2.558} \\
    \bottomrule
    \end{tabular}}
  \label{tab: Quantitative_compare}
\end{table}

\textbf{Datasets and Metrics.} To evaluate the performance of the proposed method, we conduct experiments on the test sets of two paired datasets that contain paired low-light and normal-light images, including LOL~\cite{RetinexNet} and LSRW~\cite{R2RNet}, as well as three real-world unpaired benchmarks that contain low-light images only, including DICM~\cite{DICM}, NPE~\cite{NPE}, and VV~\cite{VV}. For paired datasets, we adopt two distortion metrics PSNR and SSIM~\cite{SSIM}, and a full-reference perceptual metric LPIPS~\cite{LPIPS} for evaluation. For unpaired datasets, we use two non-reference perceptual metrics NIQE~\cite{NIQE} and PI~\cite{PI} to measure the visual quality.

\subsection{Comparison with Existing Methods}\label{subsec:comparisons}
\textbf{Comparison Methods.} We compare our method with four categories of existing LLIE methods: 1) traditional methods including LIME~\cite{LIME}, SDDLLE~\cite{SDDLLE}, BrainRetinex~\cite{BrainRetinex}, and CDEF~\cite{CDEF}, 2) supervised methods including RetinexNet~\cite{RetinexNet}, KinD++~\cite{KinD++}, LCDPNet~\cite{LCDPNet}, URetinexNet~\cite{Uretinex-net}, SMG~\cite{SMG}, PyDiff~\cite{PyDiff}, and GSAD~\cite{GSAD}, 3) semi-supervised methods DRBN~\cite{DRBN} and BL~\cite{BL}, 4) unsupervised methods including Zero-DCE~\cite{Zero-DCE}, EnlightenGAN~\cite{EnlightenGAN}, RUAS~\cite{RUAS}, SCI~\cite{SCI}, GDP~\cite{GDP}, PairLIE~\cite{PairLIE}, and NeRCo~\cite{NeRCo}. Note that supervised methods are trained on the LOL training set, and the reported performance of GDP and our method are the mean values for five times evaluation. 
\begin{figure}[!t]
    \centering
    \includegraphics[width=\linewidth]{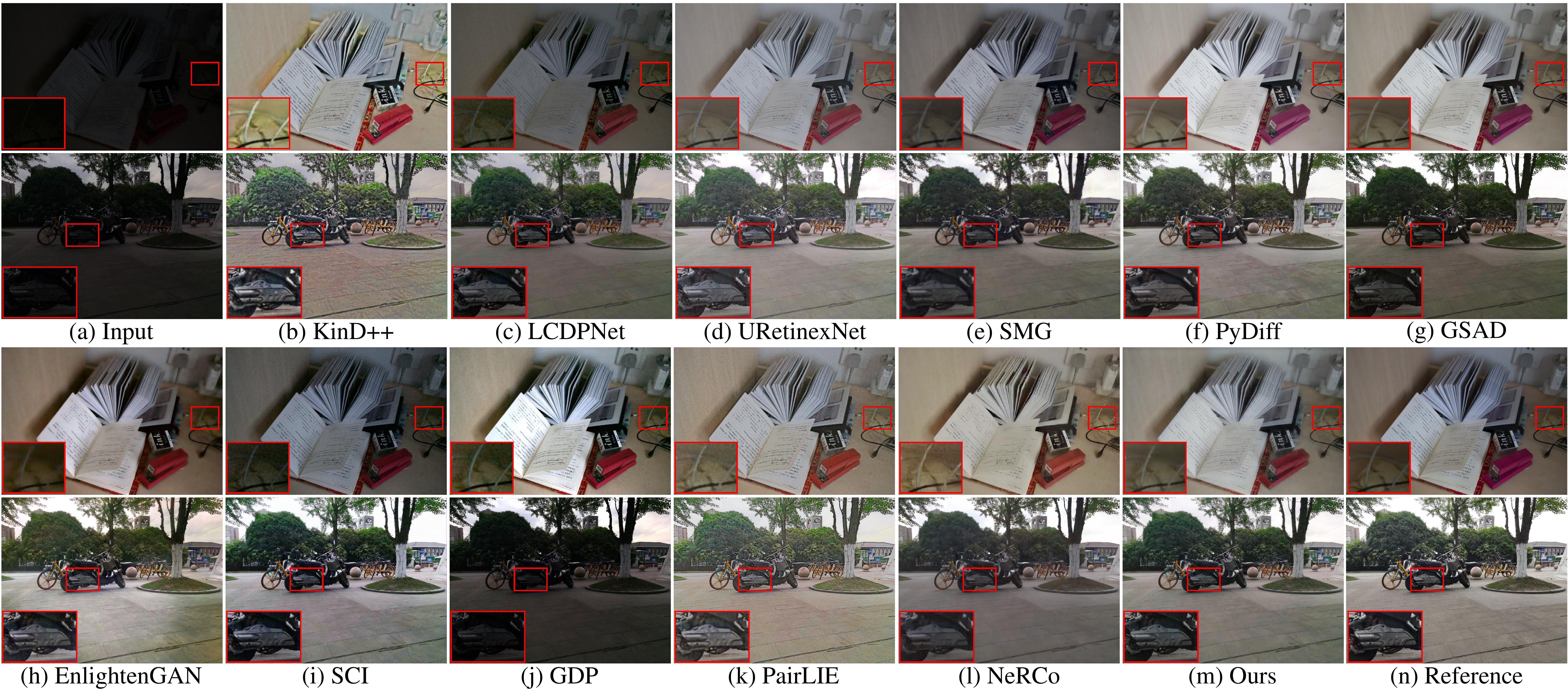}
    \caption{Qualitative comparison of our method and competitive methods on the LOL~\cite{RetinexNet} and LSRW~\cite{R2RNet} test sets. Best viewed by zooming in.}
    \label{fig: paired_comparison}
\end{figure}
\begin{figure}[!t]
    \centering
    \includegraphics[width=\linewidth]{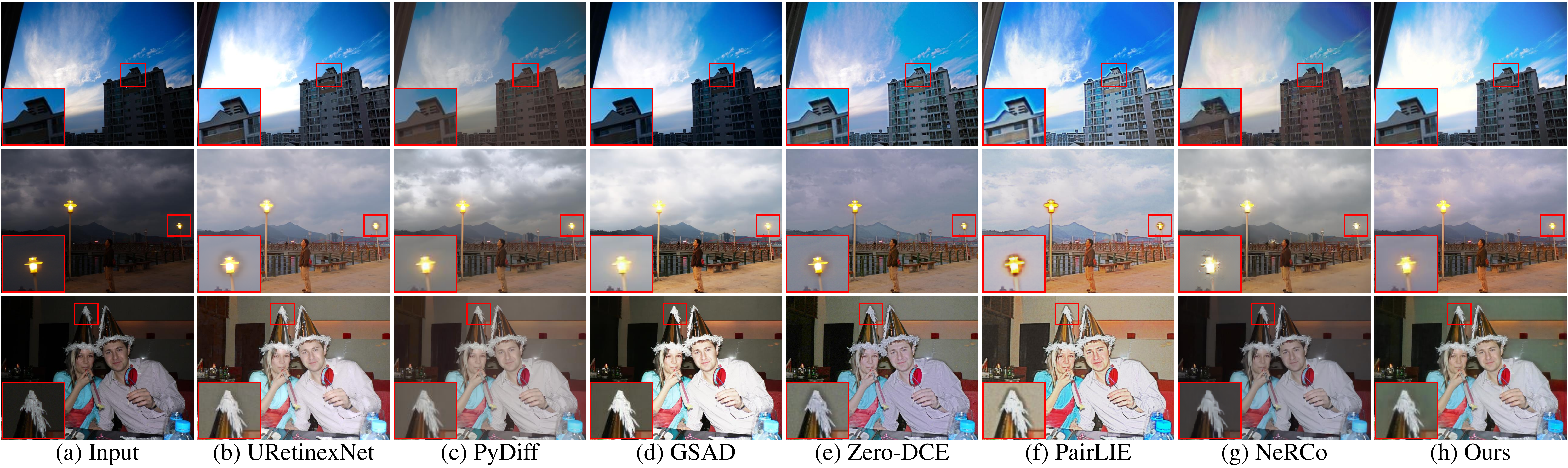}
    \caption{Qualitative comparison of our method and competitive methods on the DICM~\cite{DICM}, NPE~\cite{NPE}, and VV~\cite{VV} datasets. Best viewed by zooming in.}
    \label{fig: unpaired_comparison}
\end{figure}

\textbf{Quantitative Comparison.} We first compare the proposed method with all comparison methods on the LOL~\cite{RetinexNet} and LSRW~\cite{R2RNet} test sets. As shown in Table~\ref{tab: Quantitative_compare}, our LightenDiffusion outperforms all unsupervised competitors on both two benchmarks. The reason we cannot surpass supervised approaches on the LOL dataset is that they are typically trained on it and can therefore achieve satisfactory performance. However, our method outperforms supervised methods on the LSRW dataset, achieving the highest PSNR and SSIM with slightly inferior in terms of LPIPS. To further validate the effectiveness of our method, we also compare the proposed LightenDiffusion with comparison methods on three unpaired benchmarks DICM~\cite{DICM}, NPE~\cite{NPE}, and VV~\cite{VV}. As shown in Table~\ref{tab: Quantitative_compare}, unsupervised methods present better generalization ability than supervised ones on these unseen datasets, where our method obtains the best results on all three datasets. It indicates that our method is able to generate visually satisfactory images and can generalize well to various scenes.

\textbf{Qualitative Comparison.} We present visual comparisons of our method and competitive methods on the paired datasets in Fig.~\ref{fig: paired_comparison}, where the images in rows 1-2 are selected from LOL~\cite{RetinexNet} and LSRW~\cite{R2RNet} test sets, respectively. We can see that previous methods yield results with underexposure, color distortion, or noise amplification, while our method properly improves global and local contrast, reconstructs sharper details, and suppresses noise, resulting in visually pleasing results.  We also provide results on the unpaired benchmarks in Fig.~\ref{fig: unpaired_comparison}, where the images in rows 1-3 are selected from DICM~\cite{DICM}, NPE~\cite{NPE}, and VV~\cite{VV} datasets, respectively. Previous methods fail to generalize well to these scenes, especially in row 2, where some methods present artifacts around the light or produce overexposed results. In contrast, our method presents correct exposure and vivid color, which proves the superiority of our generalization ability. 

\subsection{Low-Light Face Detection}\label{subsec:face_detection}
\begin{figure}[!t]
    \centering
    \includegraphics[width=\linewidth]{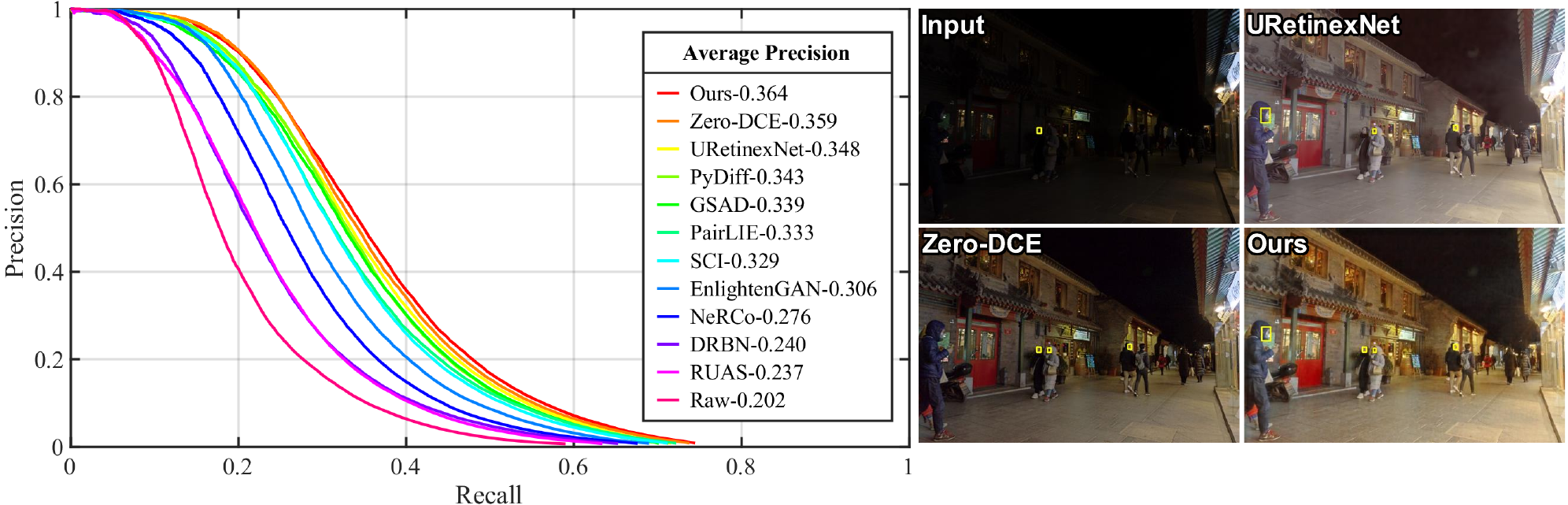}
    \caption{Comparison of low-light face detection results on the DARK FACE dataset~\cite{DarkFace}.}
    \label{fig: face_detection}
\end{figure}

In this section, we conduct experiments on the DARK FACE dataset~\cite{DarkFace}, which consists of 6,000 images captured under weakly illuminated conditions with annotated labels for evaluation, to investigate the impact of LLIE methods as a pre-processing step in improving the low-light face detection task. Following~\cite{Zero-DCE, SCI, DiffLL}, we employ our method and 10 competitive LLIE methods to restore the images, followed by the well-known detector RetinaFace~\cite{retinaface} for evaluation under the IoU threshold of 0.3 to depict the precision-recall (P-R) curves and calculate the average precision (AP). As illustrated in Fig.~\ref{fig: face_detection}, our method effectively improves the precision of RetinaFace from 20.2\% to 36.4\% compared to the raw images without enhancement and outperforms other methods in the high recall area, which reveals the potential practical values of our method.

\subsection{Ablation Study}\label{subsec:ablations}
In this section, we conduct a series of ablation studies to validate the impact of different component choices. We use the implementation details described in Sec.~\ref{subsec:experimental_settings} for training and quantitative results on the LOL~\cite{RetinexNet} and DICM~\cite{DICM} datasets are illustrated in Table~\ref{tab:ablation_study}. Detailed settings are discussed below.
\begin{table}[!t]
   \centering
    \caption{Quantitative results of ablation studies. The results using default settings are \underline{underlined}. `w/o' denotes without and `Time' denotes the inference speed when performing inference on RTX 2080Ti for an image with 400$\times$600$\times$3 resolution.}
    \resizebox{0.7\linewidth}{!}{
    \begin{tabular}{r|l|ccc|cc|c}
    \toprule
    \multirow{2}[4]{*}{} & \multirow{2}[4]{*}{Method} & \multicolumn{3}{c|}{LOL~\cite{RetinexNet}} & \multicolumn{2}{c|}{DICM~\cite{DICM}} & \multirow{2}[4]{*}{Time (s) $\downarrow$ } \\
    \cmidrule{3-7} & & PSNR $\uparrow$ & SSIM $\uparrow$ & LPIPS $\downarrow$ & NIQE$\downarrow$ & PI$\downarrow$ &  \\
    \midrule
    1)    & $k=0$ (Image Space) & 17.054 & 0.715 & 0.372 & 4.519 & 4.377 & 4.733 \\
    2)    & $k=1$ (Latent Space) & 19.228 & 0.728 & 0.355 & 4.101 & 3.457 & 0.872 \\
    3)    & $k=2$ (Latent Space) & 20.097 & 0.798 & 0.210 & 4.021 & 3.402 & 0.411 \\
    4)    & $k=4$ (Latent Space) & 20.104 & 0.785 & 0.195 & 3.906 & 3.332 & 0.256 \\
    \midrule
    5)    & RetinexNet~\cite{RetinexNet} & 16.616 & 0.563 & 0.579 & 5.859 & 6.056 & 0.296 \\
    6)    & URetinexNet~\cite{Uretinex-net} & 17.916 & 0.703 & 0.391 & 4.371 & 4.561 & 0.293 \\
    7)    & PairLIE~\cite{PairLIE} & 17.089 & 0.605 & 0.568 & 6.017 & 6.349 & 0.295 \\
    \midrule 
    8)    & w/o $\mathcal{L}_{scc}$ $(S=20)$ & 19.184 & 0.785 & 0.213 & 4.045 & 3.408 & 0.314 \\
    9)    & w/o $\mathcal{L}_{scc}$ $(S=50)$ & 19.473 & 0.791 & 0.209 & 3.998 & 3.392 & 0.687 \\
    10)   & w/o $\mathcal{L}_{scc}$ $(S=100)$ & 20.255 & 0.801 & 0.209 & 3.831 & 3.228 & 1.208 \\
    11)   & Default & \underline{20.453} & \underline{0.803} & \underline{0.192} & \underline{3.724} & \underline{3.144} & \underline{0.314}  \\
    \bottomrule
    \end{tabular}
    }
  \label{tab:ablation_study}
\end{table}
\begin{figure}[!b]
    \centering
    \includegraphics[width=0.8\linewidth]{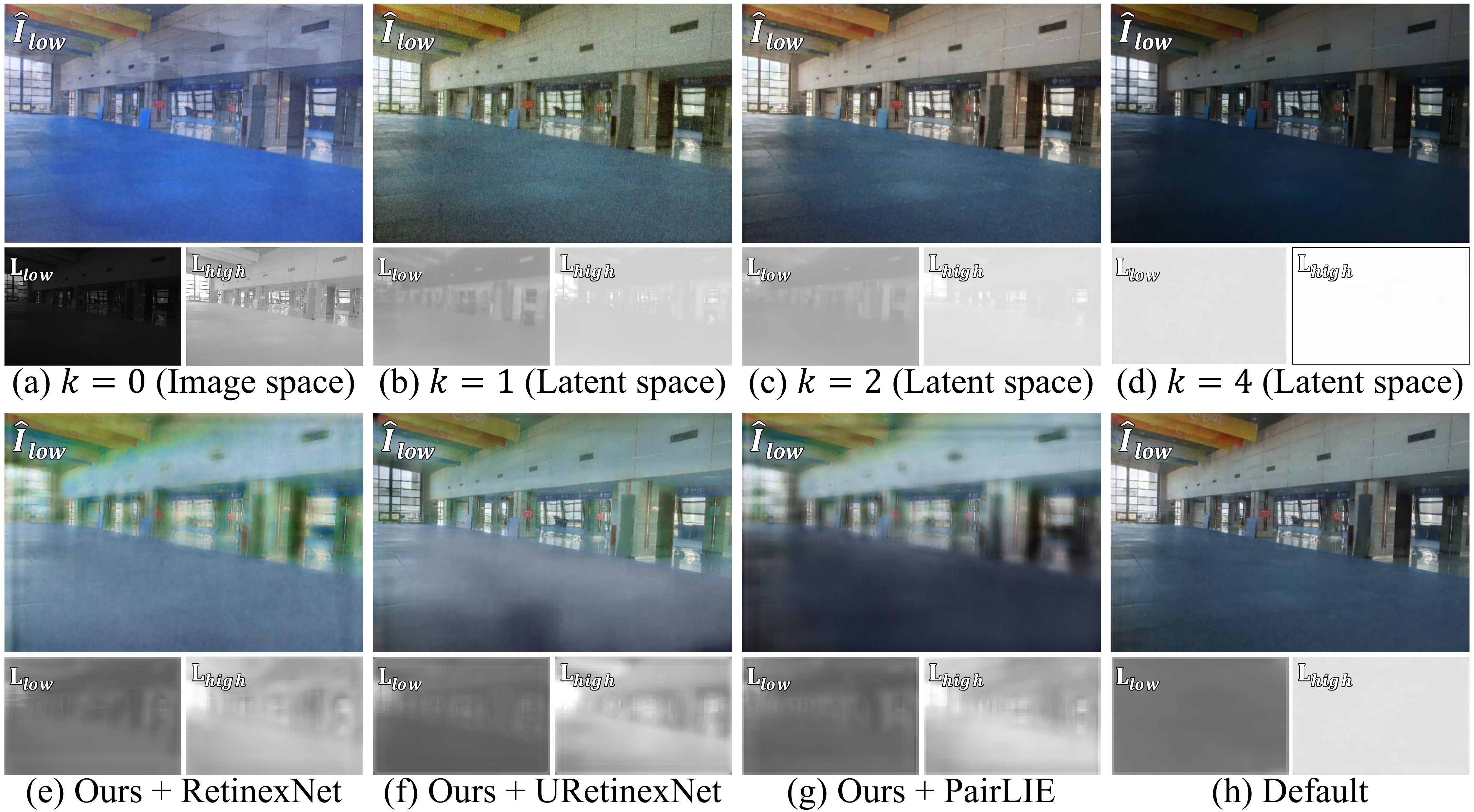}
    \caption{Visual results of the ablation study about our employed latent-Retinex decomposition strategy and the proposed content-transfer decomposition network. The first row shows the restored results with different settings, and the second row presents estimated illumination maps of low/normal-light images.}
    \label{fig: ablation_comparison_1}
\end{figure}

\textbf{Latent Space v.s. Image Space.} To validate the effectiveness of our latent-Retinex decomposition strategy, we conduct experiments by performing the decomposition in image space, i.e., $k=0$, and in various scales of latent space, i.e., $k\in[1,4]$. As shown in Fig.~\ref{fig: ablation_comparison_1}(a), it is difficult to achieve satisfactory decomposition in the image space, where the illumination map would exhibit certain content information thus making the restored image present artifacts. Conversely, as shown in Fig.~\ref{fig: ablation_comparison_1}(b)-(d), performing decomposition in the latent space can yield illumination maps that represent only the lighting conditions, which facilitates the diffusion model to generate restored images with visual fidelity. Moreover, as reported in rows 1-4 of Table~\ref{tab:ablation_study}, increasing $k$ improves the overall performance and inference speed, while showing slight performance degradation at $k=4$ due to the substantial reduction in feature information richness, which adversely affects the generative ability of the diffusion model. For a trade-off between performance and efficiency, we choose $k=3$ as the default setting.

\textbf{Retinex Decomposition Network.} To validate the effectiveness of our proposed CTDN, we replace it with the decomposition network of three previous Retinex-based methods, including RetinexNet~\cite{RetinexNet}, URetinexNet~\cite{Uretinex-net}, and PairLIE~\cite{PairLIE}, to estimate the reflectance and illumination maps. As shown in Fig.~\ref{fig: ablation_comparison_1}(e)-(g), previous decomposition networks are unable to obtain content-free illumination maps, resulting in the restored results with blurry details and artifacts. In contrast, our method benefits from the well-designed network architecture of CTDN that enables the generation of content-rich reflectance maps and content-free illumination maps, resulting in remarkable performance superiority in comparison, as reported in Table~\ref{tab:ablation_study}.
\begin{figure}[!t]
    \centering
    \includegraphics[width=0.95\linewidth]{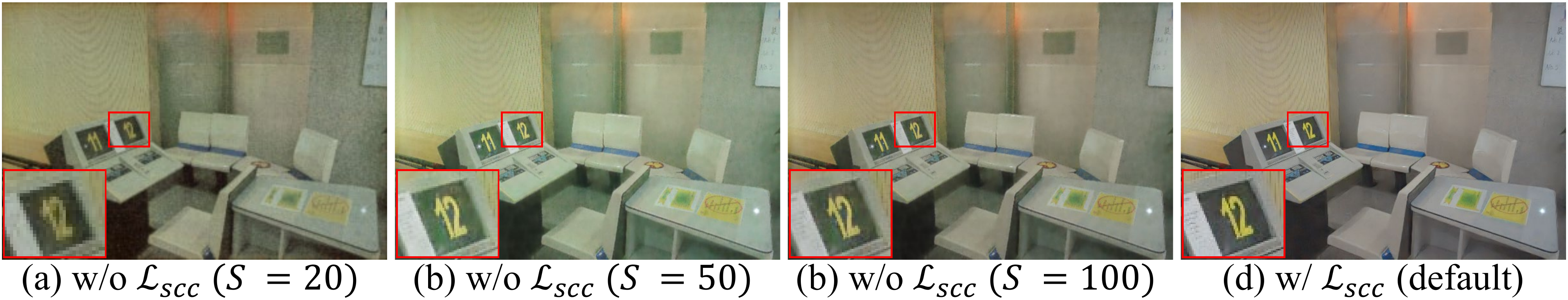}
    \caption{Visual results of the ablation study about our proposed $\mathcal{L}_{scc}$.}
    \label{fig: ablation_comparison_2}
\end{figure}

\textbf{Loss Function.} To validate the effectiveness of the proposed self-constrained consistency loss $\mathcal{L}_{scc}$, we conduct an experiment to remove it from the object function utilized to optimize the diffusion model. As reported in row 8 of Table~\ref{tab:ablation_study}, removing $\mathcal{L}_{scc}$ results in decreased overall performance. Moreover, we increase the sampling step $S$ to 50 and 100 to evaluate the performance of the diffusion model trained with vanilla diffusion loss, i.e., Eq.(\ref{eq:5}), since the quality of generated results from diffusion models would improve with increasing $S$~\cite{ddim}, as shown in Fig.~\ref{fig: ablation_comparison_2}. Compared to the default setting in row 11, while increasing the sampling step size to $S=100$ yields comparable performance to the model trained with $\mathcal{L}_{scc}$, it results in almost 4 times slower inference speed, which proves our loss can facilitate the model to achieve efficient and robust restoration.

\section{Conclusion}\label{sec:conclusion}
We have presented LightenDiffusion, a diffusion-based framework that incorporates Retinex theory with diffusion models for unsupervised LLIE. Technically, we propose a content-transfer decomposition network that performs decomposition within the latent space to obtain content-rich reflectance maps and content-free illumination maps to facilitate subsequently unsupervised restoration. The reflectance map of the low-light image and the illumination map of the normal-light image captured in different scenes serve as inputs to the diffusion model for training. Moreover, we propose a self-constrained consistency loss to further constrain the restored result to have the same inherent content information as the low-light input. Experimental results show that our method outperforms state-of-the-art competitors both quantitatively and visually. 

\textbf{Acknowledgements.} This work was supported in part by the National Natural Science Foundation of China (Nos.62372091, 62301310), the Sichuan Science and Technology Program of China (Nos.2023NSFSC0462, 2024NSFSC0944), and the funding from Sichuan University (No.2024SCU12060).
%
%
\bibliographystyle{splncs04}
\bibliography{main}
\end{document}